\documentclass[letterpaper]{article} 
\usepackage{aaai2026} 
\usepackage{times}  
\usepackage{helvet}  
\usepackage{courier}  
\usepackage[hyphens]{url}  
\usepackage{graphicx} 
\usepackage{natbib}  
\usepackage{caption} 
\usepackage{algorithm}

\usepackage{booktabs}
\usepackage{amsmath}
\usepackage{algpseudocode}
\usepackage{array}
\usepackage{multirow}

\usepackage{newfloat}
\usepackage{listings}
\DeclareCaptionStyle{ruled}{labelfont=normalfont,labelsep=colon,strut=off} 
\floatstyle{ruled}
\newfloat{listing}{tb}{lst}{}
\floatname{listing}{Listing}
\title{An Overall Real-Time Mechanism for Classification and Quality Evaluation of Rice}

\author {
    Wanke Xia\textsuperscript{\rm 1,2}\equalcontrib,
    Ruoxin Peng\textsuperscript{\rm 1}\equalcontrib,
    Haoqi Chu\textsuperscript{\rm 1},
    Xinlei Zhu\textsuperscript{\rm 1},
    Zhiyu Yang\textsuperscript{\rm 1},
    Yiting Zhao\textsuperscript{\rm 2},
    Lili Yang\textsuperscript{\rm 1}\thanks{Corresponding Author.}
}
\affiliations {
    \textsuperscript{\rm 1} China Agricultural University, \textsuperscript{\rm 2} Tsinghua University \\
    llyang@cau.edu.cn
}

\usepackage{bibentry}

\begin{document}

\maketitle

\begin{abstract}
Rice is one of the most widely cultivated crops globally and has been developed into numerous varieties. The quality of rice during cultivation is primarily determined by its cultivar and characteristics. Traditionally, rice classification and quality assessment rely on manual visual inspection, a process that is both time-consuming and prone to errors. However, with advancements in machine vision technology, automating rice classification and quality evaluation based on its cultivar and characteristics has become increasingly feasible, enhancing both accuracy and efficiency. This study proposes a real-time evaluation mechanism for comprehensive rice grain assessment, integrating a one-stage object detection approach, a deep convolutional neural network, and traditional machine learning techniques. The proposed framework enables rice variety identification, grain completeness grading, and grain chalkiness evaluation. The rice grain dataset used in this study comprises approximately 20,000 images from six widely cultivated rice varieties in China. Experimental results demonstrate that the proposed mechanism achieves a mean Average Precision (mAP) of 99.14\% in the object detection task and an accuracy of 97.89\% in the classification task. Furthermore, the framework attains an average accuracy of 97.56\% in grain completeness grading within the same rice variety, contributing to an effective quality evaluation system.
\end{abstract}

\begin{links}
    \link{Datasets}{https://huggingface.co/datasets/xwk25/RiceCC}
\end{links}

\section{Introduction}

Rice is one of the most important crops world-wide and serves as a staple food for more than half of the global population~\cite{fairhurst2002rice}. Its quality has a significant impact on dietary health. Additionally, the cultivar and characteristics of rice are key factors influencing its market value and selling price~\cite{charoenthaikij2021quality}. Therefore, the ability to assess rice quality rapidly and accurately is of critical importance. Rice quality is determined by a wide range of indicators. For instance, type quality is a fundamental factor in assessing the nutritional value of rice, while processing quality further influences its market competitiveness based on type quality~\cite{verma2018nutritional}. Currently, manual sensory evaluation remains the most widely used method for rice quality assessment~\cite{simonelli2017chemical}. However, this approach is susceptible to variations in lighting conditions, human eyesight, emotions, and other subjective factors, leading to slow identification speeds and an inability to meet the demands of rapid and objective evaluation.



With computer vision advancing rapidly, machine learning is increasingly applied in agriculture, enabling breakthroughs in crop detection~\cite{zhang2024gait}. It excels at efficient data analysis and processing, proving valuable for rice quality assessment~\cite{robert2020cascade,koklu2021classification,ahad2023comparison}. 
Several studies have demonstrated its effectiveness in this field. 
For example, ~\cite{moses2022deep} achieved single-class accuracies of 98.33\%, 96.51\%, 95.45\%, 100\%, 100\%, 99.26\%, and 98.72\% for healthy, fully chalky, chalky discolored, semi-chalky, broken, discolored, and normal damage classes, respectively, attaining an overall classification accuracy of 98.37\% using the EfficientNet-B0 architecture. 
~\cite{din2024ricenet} developed the RiceNet model based on deep convolutional neural networks, using a dataset of over 4,700 images to classify five rice varieties, achieving an accuracy of 94\%. 
~\cite{lin2018deep} proposed a machine vision system based on deep convolutional neural networks for rice classification, obtaining an accuracy of 95.5\%. 
~\cite{zareiforoush2016qualitative} applied heuristic classification methods combined with computer vision to classify four rice types, achieving the highest classification accuracy of 98.72\% using an artificial neural network topology. 
~\cite{kurtulmucs2015discriminating} introduced a cost-effective method based on computer vision and machine learning, reporting an overall accuracy of 99.24\% with the best predictive model. 

In the field of rice quality evaluation, research utilizing computer vision in accordance with national or international standards remains limited. 
For instance, ~\cite{zareiforoush2016qualitative} provided only a qualitative grading of rice quality without employing quantitative criteria for comprehensive quality assessment and classification. 
Similarly, ~\cite{moses2022deep} categorized full and semi-chalky rice but did not measure chalkiness based on national standards. 

\begin{table*}[t]
\centering
\small
\caption{The basic information of each type of rice in our manually-collected dataset.}
\label{tab:1}
\resizebox{\textwidth}{!}{
\begin{tabular}{cccccc}
\toprule
\textbf{Type Name} & \textbf{Abbr} & \textbf{Rice Variety} & \textbf{Avg Length (mm)} & \textbf{Avg Width (mm)} & \textbf{Phenotypes} \\
\midrule
Guangdong Simiao Rice & GD  & Indica Rice & 6.74 & 1.74 & Low or semi opacity, slim figure \\
Northeastern Glutinous Rice & NM  & Glutinous Rice & 4.45 & 2.86 & Wax white color, high opacity, plump figure \\
Wuchang Rice & WC  & Indica Rice & 6.63 & 2.44 & High opacity, plump figure \\
Panjin Crab Field Rice & PJX  & Japonica Rice & 4.82 & 2.83 & High opacity, plump figure \\
Wannian Gong Rice & WN  & Indica Rice & 6.81 & 2.20 & Low or semi opacity, slim figure \\
Yanbian Rice & YB  & Japonica Rice & 4.59 & 2.62 & Low or semi opacity, plump figure \\
\bottomrule
\end{tabular}
}
\end{table*}

To address prior study challenges, we propose an overall real-time mechanism for classification and quality evaluation of rice.
Our primary contributions are as follows:

\begin{itemize}
    \item An improved one-stage object detection approach is employed to identify different rice types in mixed-species samples and analyze their distribution patterns.
    \item An improved CNN is utilized to assess the completeness of six rice varieties, achieving a 2.0\% increase in accuracy compared to other classification models.
    \item The K-means clustering algorithm is applied to quantify the chalky area in rice grains, providing a precise measurement for quality evaluation.
    \item A dataset with 20,000 images spanning 6 major cultivated rice varieties in China is manually obtained, thereby furnishing a high-caliber data foundation.
\end{itemize}

\section{Data Acquisition}\label{data}
We have primarily collected an image dataset comprising six widely cultivated rice varieties in China, with their basic characteristics detailed in Table \ref{tab:1}. Phenotypic images of individual rice grains were captured using an industrial camera, and each grain was manually annotated based on its completeness and chalkiness. The hardware setups for data acquisition include a Sony CMOS sensor (providing real-time imaging at $1920\times1080$P resolution), an AOSVI T2-HD228S body microscope, a Philips monitor, and other peripheral devices. The image acquisition was conducted as follows. First, rice grains were evenly spread on a carrier table with a monochrome background. Next, optical parameters were adjusted to obtain clear, high-magnification images, after which the industrial camera was used to capture phenotypic images. Finally, the acquired images underwent manual screening and labeling to ensure data quality. Following this procedure, we collected approximately 20,000 images of Chinese rice grains against a solid-color background. These images were manually filtered and labeled to construct a dataset for rice object detection and completeness assessment. To enhance model's robustness, each rice variety dataset included both intact and broken grains in a ratio of approximately 10:1. Also, to investigate the effect of varying optical conditions on rice chalkiness detection, we selected 10 grains from each variety and captured their images under 5 different lighting environments, yielding a total of 300 images dedicated to the rice chalkiness experiment. Detailed image processing procedures applied in later experiments can be found at Appendix \ref{appendix.A}.


\section{Methods}\label{methods}

\begin{figure*}[hbt!]
    \centering
    \small
    \includegraphics[width=1\linewidth]{images/fig3.pdf}
    \caption{Framework of our improved YOLO-v5.}
    \label{fig:3}
\end{figure*}



\subsection{The One-Stage Model for Object Detection}

One-stage object detection models perform detection and classification in a single forward pass without prior candidate regions. The YOLO series, a leading framework, has advanced real-time object detection with high accuracy~\cite{jocher2020ultralytics}. 
Our research enhances performance via attention to YOLO-v5. 
YOLO-v5 comprises three primary components, which are backbone, neck, and head. The backbone is responsible for extracting feature representations from the input image. The neck further processes and fuses these features. The head predicts bounding boxes, class probabilities, and objectness scores to assess object properties.
To enhance accuracy and mAP, we replace the third C3 layer with SimAM~\cite{yang2021simam} at the backbone, as shown in Figure \ref{fig:3}. 
Unlike conventional attention mechanisms that typically introduce additional parameters or complex computations to emphasize or suppress features, SimAM adopts a self-induced approach that optimizes an energy function to assess the importance of individual neurons, as shown in Algorithm \ref{alg:1}. 
This enables the derivation of an analytical solution for feature weighting without increasing the network’s parameter count.

\begin{algorithm}[H]
\small
\caption{SimAM Implementation.}
\label{alg:1}
\begin{flushleft}
\textbf{Input:} \( X \) (input feature with shape \([N, C, H, W]\)), \( \lambda \) (coefficient parameter) \\
\textbf{Output:} Attended features \\
\end{flushleft}
\begin{algorithmic}[1]
\State \(\mathbf{n} \gets X.\text{shape}[2] \odot X.\text{shape}[3] - 1\) \Comment{spatial size}
\State \(\mathbf{X}_{\text{mean}} \gets X.\text{mean}(\text{dim}=[2, 3])\) \Comment{mean of \( X \) over spatial dimensions}
\State \(\mathbf{d} \gets (X - \mathbf{X}_{\text{mean}}).\text{pow}(2)\) 
\State \(\mathbf{v} \gets \mathbf{d}.\text{sum}(\text{dim}=[2, 3]) / \mathbf{n}\) \Comment{channel variance}
\State \(\mathbf{E\_inv} \gets \mathbf{d} / (4 \cdot (\mathbf{v} + \lambda)) + 0.5\) 
\State \(\text{return } X \odot \text{sigmoid}(\mathbf{E\_inv})\) \Comment{return attended features}
\end{algorithmic}
\end{algorithm}

\subsection{The CNN Model for Completeness Evaluation}

\begin{figure}[hbt!]
    \centering
    \includegraphics[width=1\linewidth]{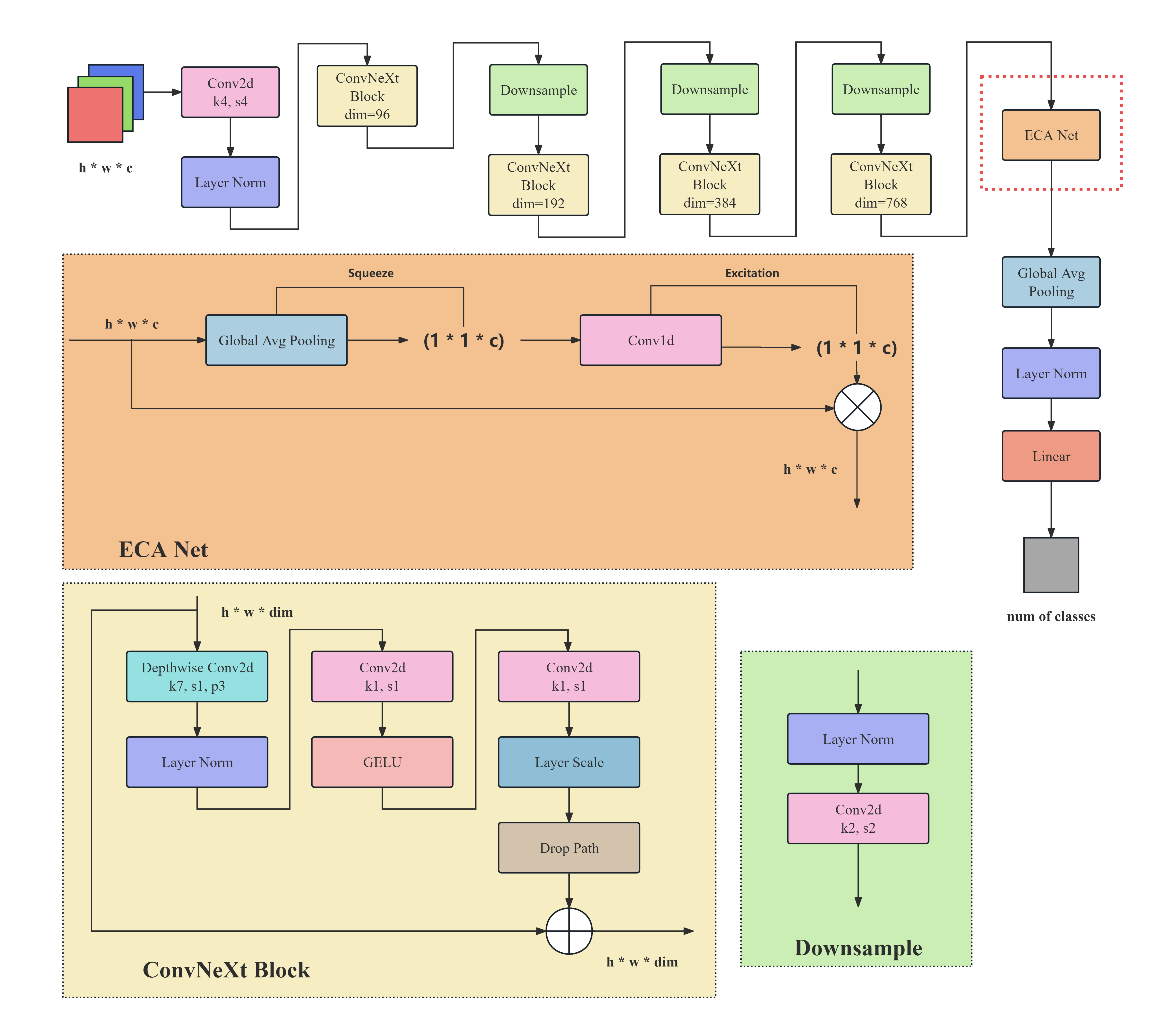}
    \caption{Framework of our improved ConvNeXt-Tiny.}
    \label{fig:4}
\end{figure}

We enhance ConvNeXt-Tiny~\cite{liu2022convnet} with attention specifically tailored for rice completeness recognition. 
The core building foundation of ConvNeXt-Tiny is the ConvNeXt Block, which consists of multiple group convolution operations and transformation layers. By stacking multiple ConvNeXt Blocks, ConvNeXt-Tiny can be constructed to meet the complexity of the task. 
In this architecture, the output of the ConvNeXt Blocks is connected to a global average pooling layer and fully connected layers, with classification ultimately performed using the softmax function.

To improve feature extraction and refine channel-wise attention, we modify ConvNeXt-Tiny by incorporating the Efficient Channel Attention (ECA)~\cite{wang2020eca} between the final ConvNeXt Block and the global average pooling layer, as shown in Figure \ref{fig:4}. 
ECA operates by computing attention weights independently for each feature map channel without considering spatial relationships between pixels or regions, allowing it to efficiently capture critical channel-wise information while avoiding computationally expensive pairwise interactions, thus well-suited for enhancing rice completeness recognition.
Additionally, to reduce training costs and develop a high-accuracy yet practical model, we initialized the network using pre-trained ConvNeXt-Tiny weights from ImageNet-1K.

\subsection{The K-means Algorithm for Chalk Recognition}

\begin{algorithm}[H]
\small
\caption{K-means Algorithm for Chalk Recognition.}
\label{alg:2}
\textbf{Input:} Image pixels $\{p_1, ..., p_N\}$, clusters $K$, convergence threshold $\epsilon$, category threshold $T$ \\
\textbf{Output:} Segmented image with chalky (black) and non-chalky (white) regions
\begin{algorithmic}[1]
\State \textbf{Initialize} $K$ cluster centers $\{\mu_1, \mu_2, ..., \mu_K\}$ randomly from image pixels
\State \textbf{Repeat}
\State \quad \textbf{Assignment step:} Assign each pixel $p_i$ to the nearest cluster $c$
\State \qquad where $c = \arg\min_j \|p_i - \mu_j\|^2$
\State \quad \textbf{Update step:} Recalculate each cluster center as the mean of assigned pixels
\State \qquad $\mu_j = \frac{1}{|C_j|} \sum_{p_i \in C_j} p_i$ where $C_j$ is the set of pixels in cluster $j$
\State \textbf{Until} Change in cluster centers is less than $\epsilon$ (convergence)

\State \textbf{Segmentation:}
\For{each pixel $p_i$ in the image}
\State Determine its cluster $j$ from the assignment step
\If{Cluster $j$ meets chalky criteria based on $T$}
\State Mark pixel as black (chalky region)
\Else
\State Mark pixel as white (non-chalky region)
\EndIf
\EndFor

\State \Return Segmented image
\end{algorithmic}
\end{algorithm}

The recognition of rice chalk involves segmenting chalky and non-chalky pixel areas within rice grains. Chalk refers to the white, opaque portion in the rice endosperm, resulting from reduced translucency caused by gaps between endosperm starch granules. In chalk recognition, the K-means algorithm (as shown in Algorithm \ref{alg:2}) is employed to assign image pixels to $K$ cluster centers for image segmentation, and the determination of $K$ is strictly based on the intrinsic spatial distribution characteristics of rice chalkiness.

Chalky areas typically aggregate in the middle region of rice grains, presenting as a single continuous opaque zone rather than discrete multiple regions. This spatial regularity indicates that the pixel set of chalky areas in rice grain images belongs to a single homogeneous cluster, while non-chalky pixels with consistent translucency and color form another dominant cluster. However, since the core goal of this study is to extract and calculate the area of the chalky region, we only need to focus on the cluster corresponding to chalky pixels. Thus, 
$K$ is set to 1, where the K-means algorithm will iteratively converge to the cluster center that best represents chalky pixels. After clustering, we extract this single cluster, calculate its total pixel count, and convert it to the actual area of the rice chalky region based on the image resolution calibration parameters.

\section{Experiments and Results}\label{exps}
All experiments were conducted on a high-performance computing system equipped with an AMD EPYC 7543 32-core processor, an Nvidia RTX 3090, and Ubuntu 20.04 LTS.
We primarily adhere to the Chinese National Standard \textit{GB/T 1354-2018 Rice} as the assessment benchmark~\cite{zhang2021overview}.
Detailed descriptions about \textit{GB/T 1354-2018 Rice} can be found at Appendix \ref{appendix.B}.

\subsection{Object Detection Experiment}


We conducted a comparative analysis between our improved one-stage model and other two-stage models. The test accuracy for each model is presented in Table \ref{tab:5}. The results indicate that two-stage models, when applied to both single-grain rice datasets and mixed-grain scenarios, fail to achieve high accuracy on the test set.
\begin{table}[htbp]
  \centering
  \small
  \caption{The test accuracy of our improved YOLO-v5 and other two-stage methods in object detection experiment.}
  \label{tab:5}
  \begin{tabular}{cc}
    \toprule
    \textbf{Model} & \textbf{Test Accuracy} \\
    \midrule
    
    Faster-RCNN (two-stage) ~\cite{ren2016faster} & 87.33\% \\
    Tridentnet (two-stage)~\cite{li2019scale} & 93.69\% \\
    YOLO-v5 (one-stage) & 95.05\% \\
    Ours (YOLO-v5 with SimAM, one-stage) & \textbf{97.89\%} \\
    \bottomrule
  \end{tabular}
\end{table}

\begin{table}[htbp]
  \centering
  \small
  \caption{The validation accuracy, validation mAP, and test accuracy of the original and our improved YOLO-v5.}
  \label{tab:3}
  \resizebox{0.48\textwidth}{!}{
  \begin{tabular}{cccc}
    \toprule
    \textbf{Model} & \textbf{Validation Accuracy} & \textbf{Validation mAP} & \textbf{Test Accuracy} \\
    \midrule
    YOLO-v5 & 97.54\% & 98.76\% & 95.05\% \\
    Ours(with SimAM) & \textbf{97.94\%} & \textbf{99.14\%} & \textbf{97.89\%} \\
    \bottomrule
  \end{tabular}
  }
\end{table}

\begin{figure}[hbt!]
    \centering
    \includegraphics[width=0.9\linewidth]{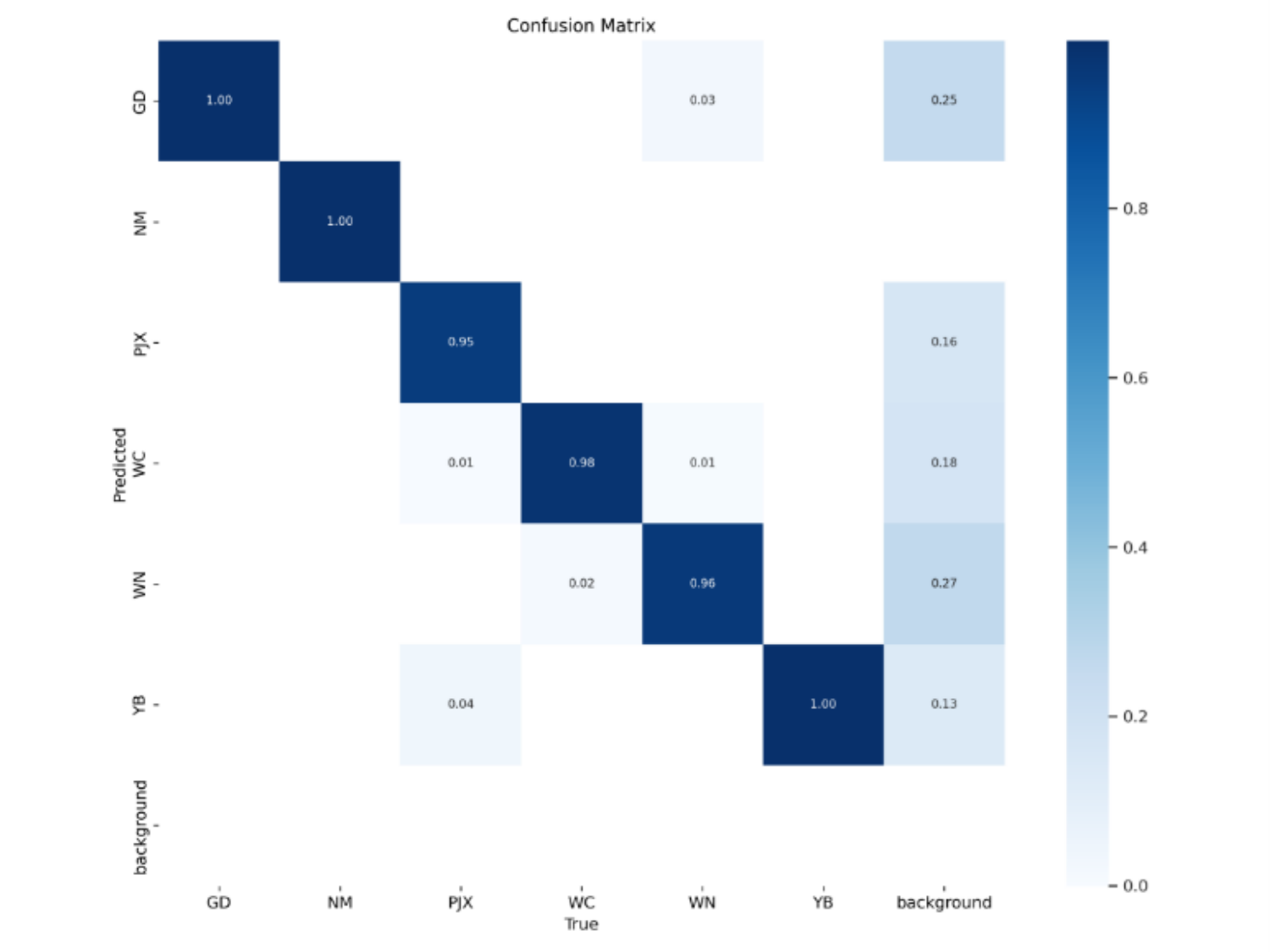}
    \caption{The confusion matrix of our improved YOLO-v5 on the validation set.}
    \label{fig:5}
\end{figure}

The confusion matrix of our improved YOLO-v5 on the validation set is presented in Figure \ref{fig:5}. 
The model achieves a precision rate of 1.0 in GD, NM, and YB. 
Also, performance for WC, WN, and PJX is slightly lower but remained consistently above 0.95, indicating favorable overall outcomes.

\begin{figure}[hbt!]
    \centering
    \includegraphics[width=0.9\linewidth]{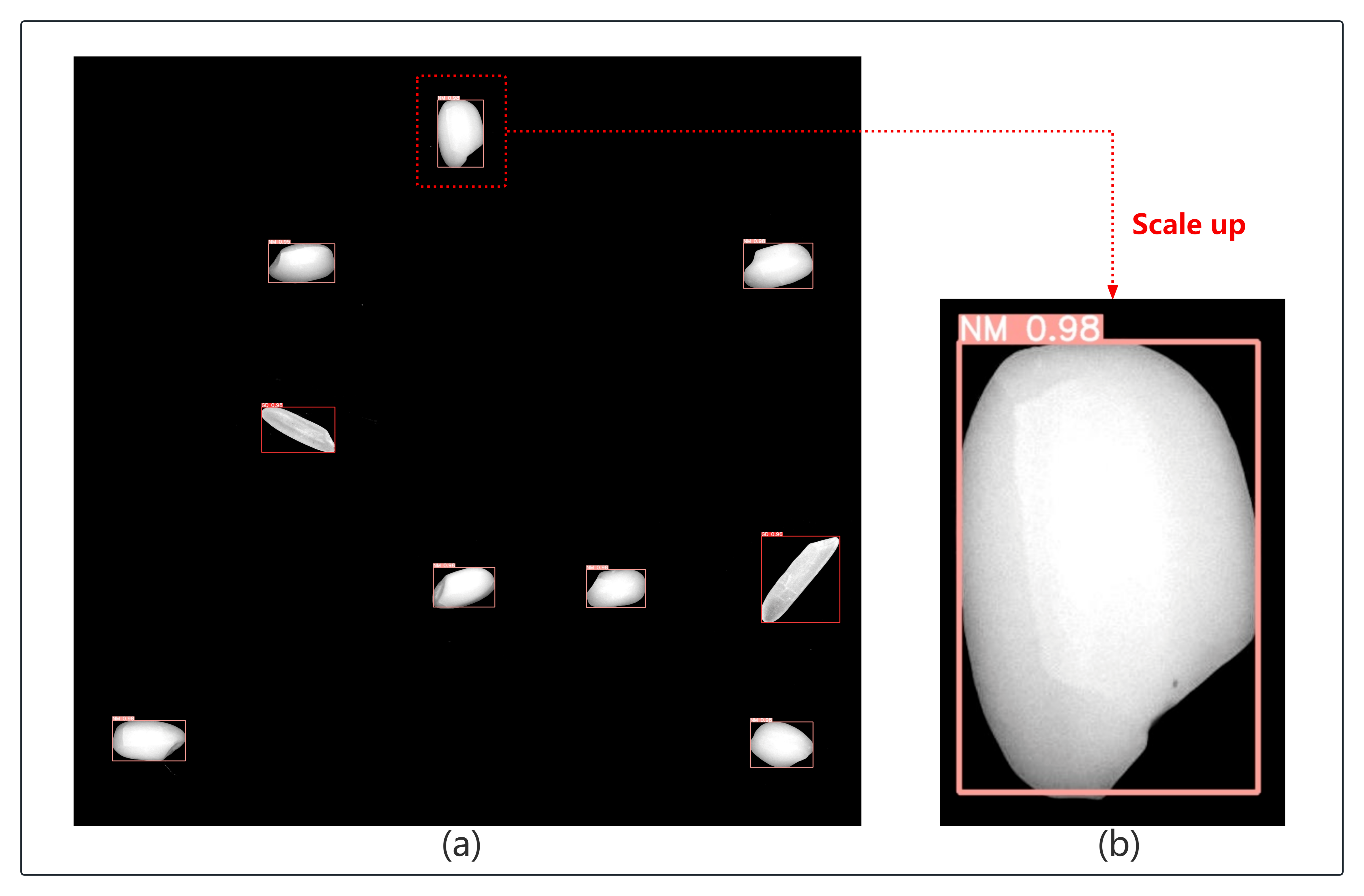}
    \caption{The example visual representation of our improved YOLO-v5 on the test set, wherein pink boxes denote NM instances, and red boxes signify GD instances. The upper-left corner of each box indicates the associated category and confidence level.}
    \label{fig:6}
\end{figure}

Example visual representation of our improved YOLO-v5 on the test set is illustrated in Figure \ref{fig:6}. 
A cluster of NM instances is intricately interwoven with a sparse presence of GD. 
Results clearly demonstrate that the anchor boxes accurately and distinctly identify both NM and the interspersed GD, achieving commendable recognition performance.

As shown in Table \ref{tab:3}, an ablation analysis is conducted between our improved YOLO-v5 and its predecessor.
It is worth noting that our improved model performs better in terms of validation accuracy and validation mAP, with increases of 0.40\% and 0.38\%, respectively.
Moreover, our improved model demonstrates an elevated accuracy of 97.89\% on the test set, significantly outperforming the original model's accuracy of 95.05\%. 
These results collectively affirm the superior precision and object detection capabilities of our improved YOLO-v5 in practical scenarios.

\begin{table*}[!t]
  \centering
  \small
  \caption{The test accuracy of various classification methods and our improved ConvNeXt-Tiny.}
  \label{tab:7}
  \resizebox{\textwidth}{!}{
  \begin{tabular}{cccccccc}
    \toprule
    \textbf{Method} & \textbf{Average Accuracy} & \textbf{GD} & \textbf{NM} & \textbf{PJX} & \textbf{WC} & \textbf{WN} & \textbf{YB} \\
    \midrule
    Decision Tree & 93.72\% & 85.47\% &	92.42\% &	97.56\% &	95.13\% &	95.13\% &	96.59\% \\
    Random Forest & 94.27\% &	87.60\% &	93.68\% &	97.88\% &	95.03\% &	94.84\% &	96.60\% \\
    SVM & 91.82\% &	87.75\% &	87.50\% &	97.72\% &	91.30\% &	91.30\% &	95.35\% \\
    Googlenet~\cite{szegedy2015going} & 89.24\% &	82.60\% &	95.65\% &	95.00\% &	95.23\% &	86.95\% &	80.00\% \\
    Alexnet~\cite{krizhevsky2012imagenet} & 94.90\% &	89.36\% &	95.74\% &	97.62\% &	\textbf{97.73\%} &	91.30\% &	97.62\% \\
    Mobilenet-v2~\cite{sandler2018mobilenetv2} & 90.45\% &	78.72\% &	91.49\% &	92.86\% &	95.45\% &	91.30\% &	92.86\% \\
    Resnet101~\cite{he2016deep} & 85.19\% &	80.85\% &	74.47\% &	90.00\% &	84.09\% &	91.30\% &	90.48\% \\
    Ours(ConvNeXt-Tiny with ECA) & \textbf{97.61\%} &	\textbf{98.80\%} &	\textbf{96.35\%} &	\textbf{98.80\%} &	96.10\% &	\textbf{95.92\%} &	\textbf{99.68\%} \\
    \bottomrule
  \end{tabular}
  }
\end{table*}

\subsection{Completeness Experiment}

We initially collected data for three levels of completeness, which are whole rice, large broken rice, and tiny broken rice. 
The dataset included measurements of their short axis, long axis, and area. 
The data quantity for each level in one rice type was varied, resulting in an approximate ratio of 2:1:1. 
Based on the definitions of broken rice outlined in \textit{GB/T 1354-2018 Rice}, we utilized various classification methods to classify and grade the completeness level of rice grains.
As shown in Table \ref{tab:7}, our improved ConvNeXt-Tiny basically achieves the best performance in both average accuracy and accuracy in each rice type. Although Alexnet shows an accuracy of 97.73\% in Wuchang rice(WC), our improved ConvNeXt-Tiny still ranks the second place with the accuracy of 96.10\%.

We have also conducted ablation experiments on the ECA module to verify its effectiveness in the framework of completeness evaluation.
As shown in Table \ref{tab:4}, our improved ConvNeXt-Tiny outweighs the original ConvNeXt-Tiny in identifying the three rice completeness grades, with an average accuracy improvement of nearly 2\%. 
Notably, significant increases in the test accuracy are observed for Guangdong Simiao rice(GD) and Wuchang rice(WC). 
The test accuracy of Guangdong Simiao rice improves from 88.65\% to 98.80\%. And the test accuracy of Wuchang rice increases from 95.44\% to 96.10\%.

\begin{table}[htbp]
  \centering
  \small
  \caption{The test accuracy of the original and our improved ConvNeXt-Tiny.}
  \label{tab:4}
  \resizebox{0.48\textwidth}{!}{
  \begin{tabular}{cccccccc}
    \toprule
    \textbf{Model} & \textbf{Average Accuracy} & \textbf{GD} & \textbf{NM} & \textbf{PJX} & \textbf{WC} & \textbf{WN} & \textbf{YB} \\
    \midrule
    
    ConvNeXt-Tiny & 95.58\% & 88.65\% & \textbf{96.35\%} & 98.28\% & 95.44\% & 95.37\% & 99.41\% \\
    Ours(with ECA) & \textbf{97.61\%} & \textbf{98.80\%} & \textbf{96.35\%} & \textbf{98.80\%} & \textbf{96.10\%} & \textbf{95.92\%} & \textbf{99.68\%} \\
    \bottomrule
  \end{tabular}
  }
\end{table}

\subsection{Chalk Experiment}
\begin{figure}[hbt!]
    \centering
    \includegraphics[width=1\linewidth]{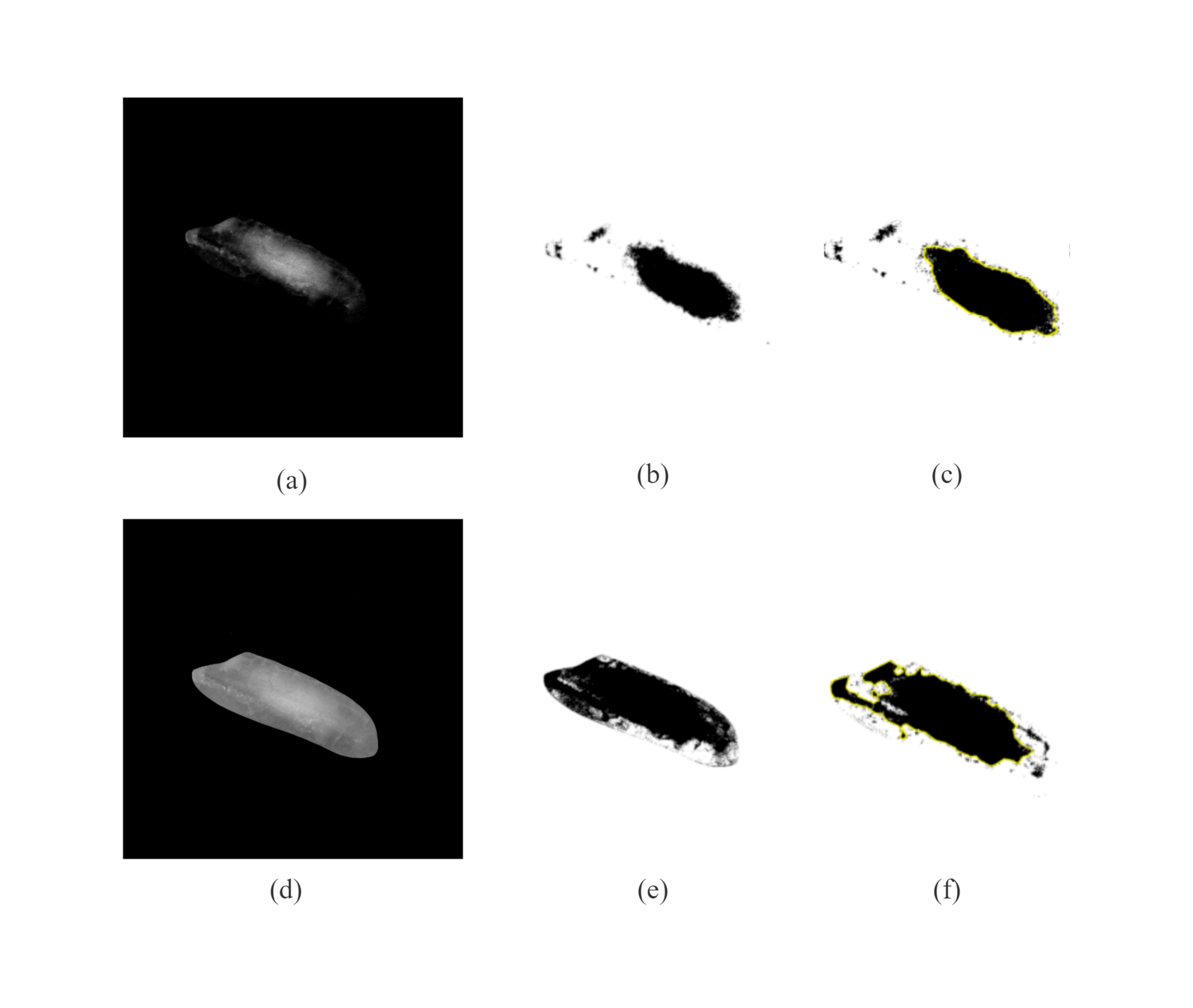}
    \caption{The processed images in chalk recognition of Guangdong Simiao Rice: (a) an original image at the brightness of level 2; (b) an image at the brightness of level 2 processed by K-means; (c) shows the theoretical area in chalk calculation surrounded by a yellow border; (d) an original image at the brightness of level 5; (e) an image at the brightness of level 5 processed by K-means; (f) shows the theoretical area in chalk calculation surrounded by a yellow border.}
    \label{fig:7}
\end{figure}

According to \textit{GB/T 1354-2018 Rice}, rice grains with a chalk content of greater than or equal to 2.0\% are classified as first-grade premium rice, implying that lower chalkiness relates to higher rice quality.
In chalk recognition, rice images are collected under 5 different luminance intensities using the same light source, then converted to grayscale for processing. 
We employed K-means to cluster the images, and by setting a category threshold, chalky regions are segmented, appearing black while non-chalky areas appear white. 
To accurately determine the chalky area, a geometric polygon fitting method was used to estimate the segmented area. The formula for area calculating is detailed in Section \ref{4.4.2}.

As shown in Figure \ref{fig:7}, the size of the chalky regions increases with higher luminance intensities. 
To ensure the reliability of the comparison benchmark, we invited several Chinese experts specializing in agriculture. 
They were asked to visually identified and determined the chalky areas of rice grains in the subset images independently, serving as the human visual assessment reference standard for subsequent analysis.
Manual comparisons revealed that the chalky area's ratio to the total rice area in images captured under luminance level one closely aligns with human visual assessments. Thus, luminance level one was selected as the standard luminance for this experiment. 
Compared to the manual visual assessment method prescribed in \textit{GB/T 1354-2018 Rice}, this automated approach is not only more precise but also significantly faster.

\section{Conclusion}
Our proposed overall mechanism, which consists of 3 primary steps, manifests the capability to accurately identify rice varieties and perform quality analyses.
First, improved YOLO-v5 is implemented to enable real-time visualization of rice positions and classifications, achieving an accuracy of 97.89\% and an mAP of 99.14\%. 
Second, enhanced ConvNeXt-Tiny is employed to evaluate the completeness grade within the same variety, yielding an average accuracy exceeding 97\%. 
Third, K-means is applied to extract the chalky regions of whole rice and calculate precise chalkiness measurements.
All experiments are conducted on our manually collected dataset with 20,000 images spanning six major cultivated rice varieties in China.
We hope that future research will focus more on establishing international standards for rice grain quality assessment and collecting multi-regional rice grain image datasets.

\bibliography{aaai2026}


\clearpage
\appendix

\section{Image Processing}\label{appendix.A}
\subsection{Image Processing for Object Detection}
\begin{figure}[hbt!]
    \centering
    \includegraphics[width=1\linewidth]{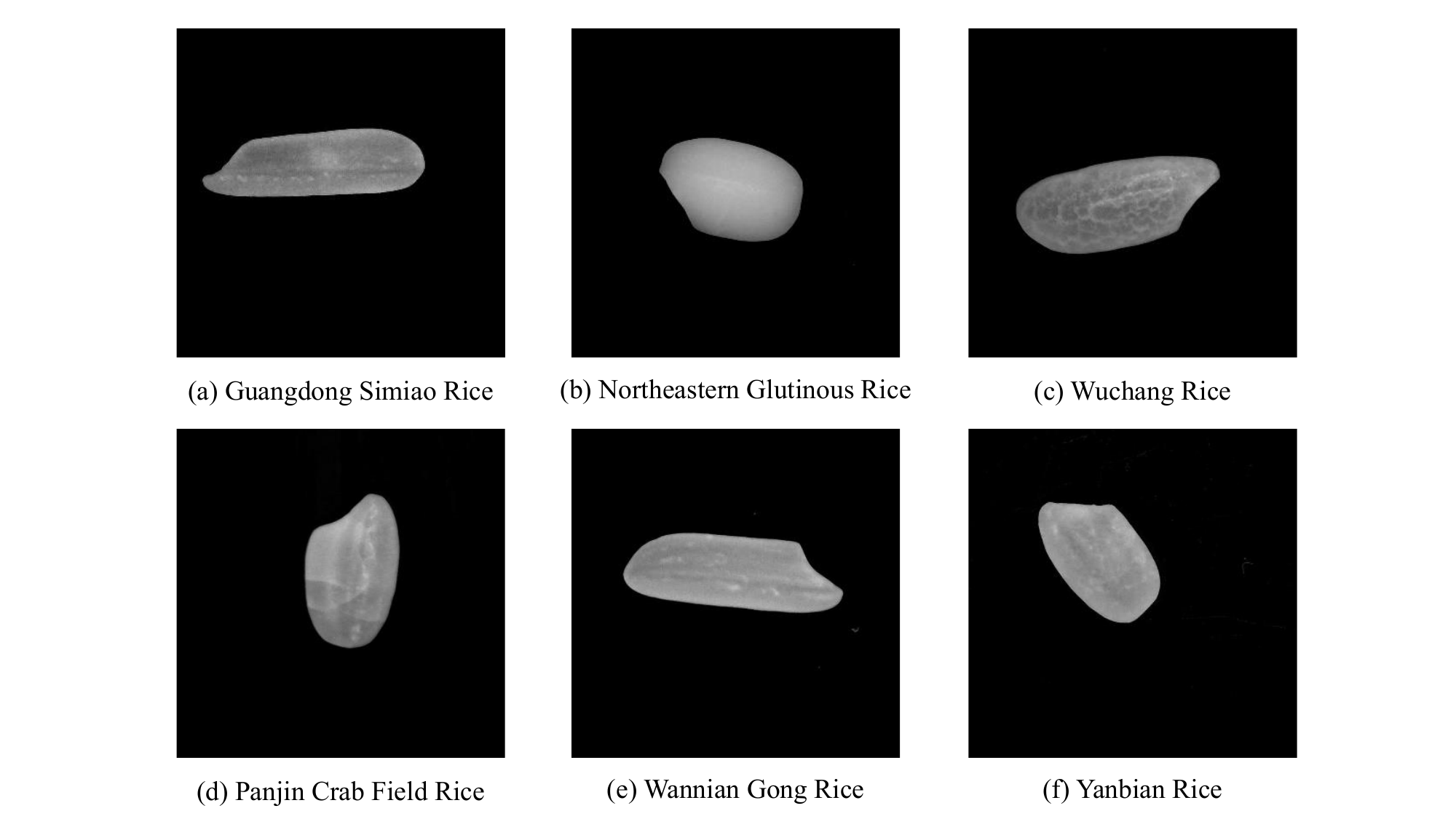}
    \caption{Example images of 6 types of Chinese rice sampled from our manually-collected dataset after grayscale processing: (a) Guangdong Simiao Rice, whose abbr is GD; (b) Northeastern Glutinous Rice,  whose abbr is NM; (c) Wuchang Rice, whose abbr is WC; (d) Panjin Crab Field Rice, whose abbr is PJX; (e) Wannian Gong Rice, whose abbr is WN; (f) Yanbian Rice, whose abbr is YB.}
    \label{fig:1}
\end{figure}

Some of the original rice images exhibited darker coloration, which obscured essential features such as translucency, chalkiness, contours, and other visual characteristics necessary for accurate observation. To enhance the visibility of these attributes, this study initially applied grayscale processing to the darker rice images. The maximum and minimum pixel values were computed, and each pixel value was remapped to a range of 0 to 255 using a lookup table, effectively replacing the original pixel values. This processing step enhanced the overall clarity, improving contour definition and sharpness, as illustrated in Figure \ref{fig:1}. Subsequently, image brightness was adjusted to further emphasize fine details and enhance contrast. Taking Guangdong Simiao rice as an example, the final brightness-adjusted image is shown in Figure \ref{fig:2} (a), demonstrating a notable improvement in brightness and visibility compared to Figure \ref{fig:1} (a). This enhancement facilitates more precise human and computational analysis.

To simulate real-world conditions where multiple rice varieties may be intermixed, we employed a randomized mixing approach to construct a test dataset for rice identification. This dataset enables the evaluation of model accuracy in classifying rice varieties under practical scenarios.

In summary, the grayscale and brightness-adjusted rice images effectively highlight key visual features such as grain outline, internal texture, chalkiness, and translucency. These enhancements ensure that the images meet the requirements for classification and recognition tasks, making them suitable as input data for relevant deep learning models.

\subsection{Image Processing for Completeness Evaluation}
For processing the broken rice image dataset, the same initial procedures were applied. In grading the completeness of rice grains, shape and contour features play a crucial role. Compared to grayscale images, binarized images more clearly delineate the shape and contours of rice grains, thereby improving the accuracy of subsequent classification algorithms. For instance, the preliminary binarized image of Guangdong Simiao rice, shown in Figure \ref{fig:2} (b), exhibits significantly more distinguishable geometric features than the original grayscale image in Figure \ref{fig:1} (a).

However, several small, unwanted white regions were present in the images. To simplify the dataset and retain only the significant and relevant areas, all white regions with a pixel area smaller than 40,000 were filled with black. Additionally, to enhance dataset quality, a median filtering algorithm was applied to remove edge noise, as illustrated in Figure \ref{fig:2} (c).

In summary, the rice images processed using the methods described in previous contexts, followed by binarization and noise reduction techniques, exhibit enhanced shape and contour clarity. These improvements contribute to more effective subsequent quality evaluation of rice grains.

\subsection{Image Processing for Chalk Recognition}
\begin{figure}[hbt!]
    \centering
    \includegraphics[width=1\linewidth]{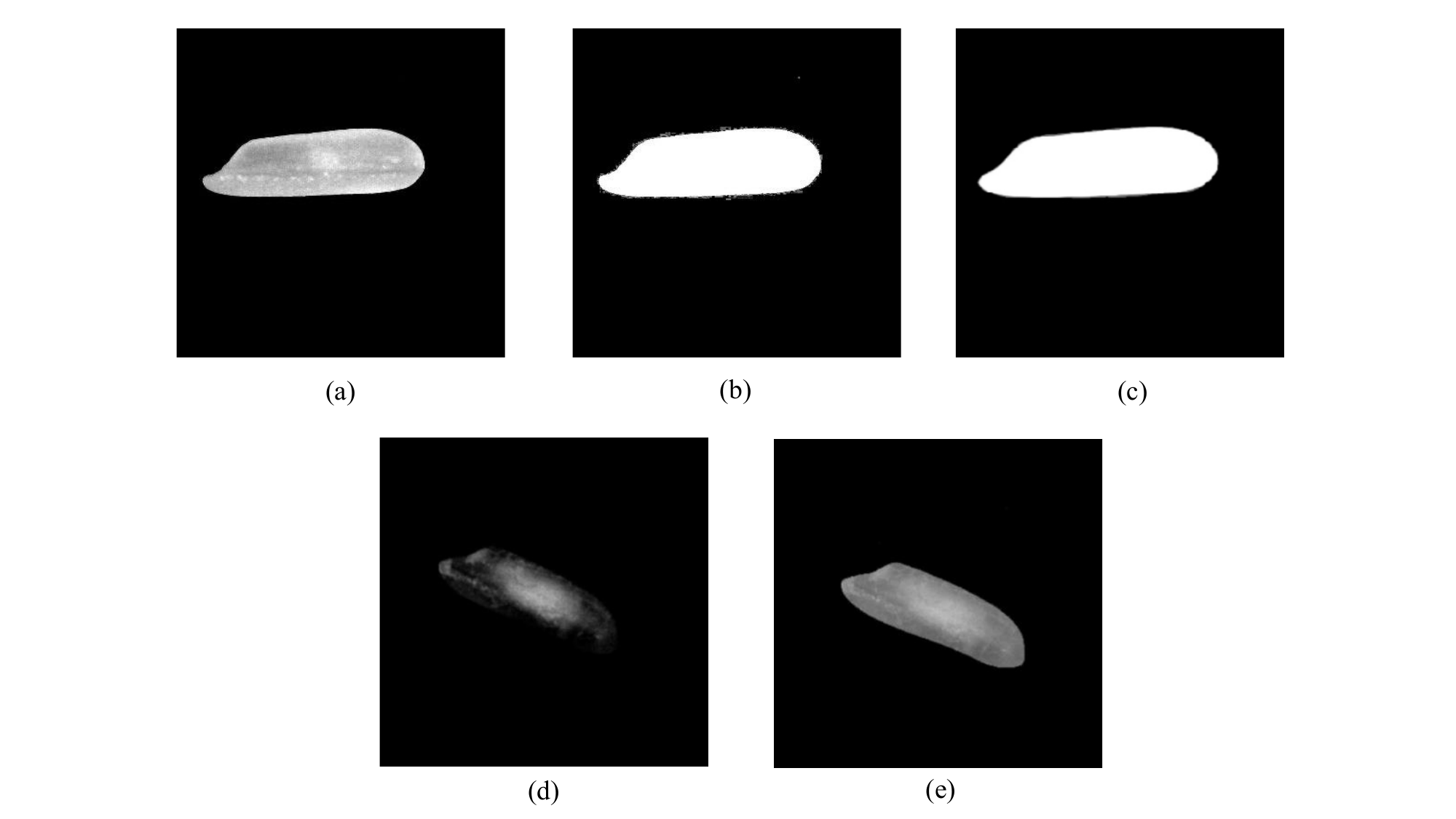}
    \caption{Example images in image processing: (a) an example image of Guangdong Simiao Rice after brightness enhancement; (b) an example image of Guangdong Simiao Rice after preliminary binarization processing; (c) an example image of Guangdong Simiao Rice after removing small domains and implementing median filtering; (d) an example image of Guangdong Simiao Rice after grayscale processing at the brightness of level 2, which is a low brightness; (e) an example image of Guangdong Simiao Rice after grayscale processing at the brightness of level 5, which is a high brightness.}
    \label{fig:2}
\end{figure}

In studies that evaluate rice chalk using machine vision, accurate chalk discrimination requires determining a specific light intensity. To investigate the impact of different light intensities on the recognition algorithm, this study established five distinct brightness levels. Using Guangdong Simiao rice as an example, Figure \ref{fig:2} (d) and Figure \ref{fig:2} (e) illustrate the results obtained at brightness levels 2 and 5, respectively.

\section{Detailed Descriptions about GB/T 1354 - 2018 Rice}\label{appendix.B}
\begin{table*}[htbp]
  \centering
  \small
  \caption{General rice quality standards of different rice varieties based on GB/T 1354 - 2018 Rice~~\cite{zhang2021overview}.}
  \label{tab:2}
  \begin{tabular}{
        >{\centering\arraybackslash}p{4.0cm}
        >{\centering\arraybackslash}p{1.4cm}
        >{\centering\arraybackslash}p{2.2cm}
        >{\centering\arraybackslash}p{2.4cm}
        >{\centering\arraybackslash}p{2.0cm}
        >{\centering\arraybackslash}p{2.0cm}
    }
    \toprule
    \textbf{Rice Variety} & \textbf{Evaluated Level} & \textbf{Broken Rice Rate} & \textbf{Small Broken Rice Rate} & \textbf{Chalk Rate} & \textbf{Admixture Rate} \\
    \midrule
    \multirow{3}{*}{Indica Rice} 
    & 1 & $\leq$15.0\% & $\leq$1.0\% & $\leq$2.0\% & \multirow{3}{*}{$\leq$5.0\%} \\
    & 2 & $\leq$20.0\% & $\leq$1.5\% & $\leq$5.0\% &  \\
    & 3 & $\leq$30.0\% & $\leq$2.0\% & $\leq$8.0\% &  \\
    \midrule
    \multirow{3}{*}{Japonica Rice} 
    & 1 & $\leq$10.0\% & $\leq$1.0\% & $\leq$2.0\% & \multirow{3}{*}{$\leq$5.0\%} \\
    & 2 & $\leq$15.0\% & $\leq$1.5\% & $\leq$4.0\% &  \\
    & 3 & $\leq$20.0\% & $\leq$2.0\% & $\leq$6.0\% &  \\
    \midrule
    \multirow{2}{*}{Glutinous Rice from Japonica branch} 
    & 1 & $\leq$10.0\% & $\leq$1.5\% & \multirow{2}{*}{/} & \multirow{2}{*}{$\leq$5.0\%} \\
    & 2 & $\leq$15.0\% & $\leq$2.0\% &  &  \\
    \midrule
    \multirow{2}{*}{Glutinous Rice from Indica branch} 
    & 1 & $\leq$15.0\% & $\leq$2.0\% & \multirow{2}{*}{/} & \multirow{2}{*}{$\leq$5.0\%} \\
    & 2 & $\leq$25.0\% & $\leq$2.5\% &  &  \\
    \bottomrule
  \end{tabular}
\end{table*}

\subsection{The Standard of Rice Completeness}
According to the classification criteria outlined in the Chinese National Standard GB/T 1354-2018 Rice, whole grains are defined as rice grains that remain intact in all parts except the embryo. Broken rice, on the other hand, refers to rice grains that are incomplete and have a length less than three-quarters of the average length of whole grains from the same batch, as retained on a 1.0 mm round-hole sieve. Within the category of broken rice, sizeable broken rice includes incomplete grains with a length less than three-quarters of the average length of whole grains from the same batch but retained on a 2.0 mm round-hole sieve. Tiny broken rice refers to fragments that pass through a 2.0 mm round-hole sieve but are retained on a 1.0 mm round-hole sieve.

To facilitate a more detailed analysis and assessment of broken rice in rice samples, we classified the rice into three categories based on the degree of completeness as defined by the standard: whole grains (with a length not less than three-quarters of the average length of whole grains from the same batch), sizeable broken rice, and tiny broken rice. Subsequently, the rice samples were labeled according to the identification criteria specified in the standard document.

For experimental validation, we used a sample of no less than 10 g of rice and conducted broken rice classification tests. The ratio of sizeable broken rice to tiny broken rice was calculated following the method prescribed in the standard, as described by Equation \ref{equ:1}.

\begin{equation}
\label{equ:1}
    X_1 = \frac{m_1}{m} \times 100\%
\end{equation}
where $X_1$ represents the broken rice rate, $m_1$ is the mass of broken rice in grams (g), and $m$ is the mass of the sample in grams (g).

Referring to the calculation method for rice's broken rice rate in the standard document, we calculate it using Equation \ref{equ:2}.

\begin{equation}
\label{equ:2}
    X_2 = \frac{m_2}{m} \times 100\%
\end{equation}
where $X_2$ represents the broken rice rate, $m_2$ is the mass of broken rice, where broken rice includes the mass of sizeable and small broken rice, measured in grams (g), and $m$ is the mass of the sample in grams (g).

\subsection{The Standard of Rice Chalk}\label{4.4.2}
According to the detection method for rice chalk specified in the Chinese National Standard GB/T 1354-2018 Rice, the procedure begins by selecting 100 intact rice grains (n0) from the suspended rice sample. From this selection, the chalky rice grains (n1) are identified and separated. Subsequently, ten chalky rice grains are randomly selected from this group for further analysis. These selected grains are placed flat and observed from the front to visually assess the chalky projection area as a percentage of the total projection area of the intact rice grains. The average of these percentages is then calculated to quantify the degree of chalkiness. The chalk content is calculated using Equation \ref{equ:3}, and the result is expressed as a percentage (\%).
\begin{equation}
\label{equ:3}
    D = W_D \times \frac{n_1}{n_0}
\end{equation}
where $D$ represents chalkiness, expressed in percentage (\%), $W_D$ represents the chalky size, also in percentage (\%), $n_1$ is the number of chalky rice grains in the sample, and $n_0$ is the total number of rice grains in the sample.

\end{document}